\documentclass{article}

\usepackage{microtype}
\usepackage{graphicx}
\usepackage{subfigure}
\usepackage{booktabs} 

\usepackage{hyperref}

\usepackage[accepted]{icml2025}

\usepackage{amsmath}
\usepackage{amssymb}
\usepackage{mathtools}
\usepackage{amsthm}

\usepackage[capitalize,noabbrev]{cleveref}

\theoremstyle{plain}
\newtheorem{theorem}{Theorem}[section]

\newtheorem{lemma}[theorem]{Lemma}

\theoremstyle{definition}

\theoremstyle{remark}

\newcommand{\tool}{\textsf{BiMarker}}
\newcommand{\kgw}{\textsf{KGW}}
\newcommand{\sweet}{\textsf{SWEET}}
\newcommand{\ewd}{\textsf{EWD}}

\usepackage[textsize=tiny]{todonotes}

\begin{document}

\twocolumn[
\icmltitle{BiMarker: Enhancing Text Watermark Detection for \\ Large Language Models with Bipolar Watermarks}


\icmlsetsymbol{equal}{*}

\begin{icmlauthorlist}
\icmlauthor{Zhuang Li}{equal,bupt}
\icmlauthor{Qiuping Yi}{bupt}
\icmlauthor{Zongcheng Ji}{pingan}
\icmlauthor{Yijian LU}{tsinghua}
\icmlauthor{Yanqi Li}{bupt}
\icmlauthor{Keyang Xiao}{bupt}
\icmlauthor{Hongliang Liang}{bupt}

\end{icmlauthorlist}

\icmlaffiliation{bupt}{Beijing University of Posts and Telecommunications, Beijing, China}
\icmlaffiliation{pingan}{Ping An Technology, Shenzhen, China}
\icmlaffiliation{tsinghua}{Tsinghua University, Beijing, China}

\icmlcorrespondingauthor{Zhuang Li}{bupt01@bupt.edu.cn}


\icmlkeywords{LLMs，Watermarking，Bipolar Watermark，Detectability}

\vskip 0.3in
]



\printAffiliationsAndNotice{}  

\begin{abstract}

The rapid growth of Large Language Models (LLMs) raises concerns about distinguishing AI-generated text from human content. Existing watermarking techniques, like \kgw, struggle with low watermark strength and stringent false-positive requirements. Our analysis reveals that current methods rely on coarse estimates of non-watermarked text, limiting watermark detectability. To address this, we propose Bipolar Watermark (\tool), which splits generated text into positive and negative poles, enhancing detection without requiring additional computational resources or knowledge of the prompt. Theoretical analysis and experimental results demonstrate \tool's effectiveness and compatibility with existing optimization techniques, providing a new optimization dimension for watermarking in LLM-generated content.

\end{abstract}
\section{Introduction}

In recent years, Large Language Models (LLMs) have revolutionized natural language processing~\cite{openai,llama,sparks}, but their widespread adoption has also raised significant concerns. 
These models can be exploited for malicious purposes, such as generating fake news, fabricating academic papers, and manipulating public opinion through social engineering and disinformation campaigns~\cite{misuse-1,misuse-2}. 
Additionally, the proliferation of synthetic data on the web complicates dataset curation, as it often lacks the quality of human-generated content and must be carefully filtered during training~\cite{misuse-3}. 
As a result, developing effective methods to reliably distinguish 
AI-generated text from human-written 
content has become a critical and urgent priority~\cite{urgent-1,urgent-2,urgent-3,urgent-4}.

\begin{figure}[t!]
\vskip 0.2in
\begin{center}
\centering\includegraphics[scale=0.25,trim=10 10 100 10,clip]{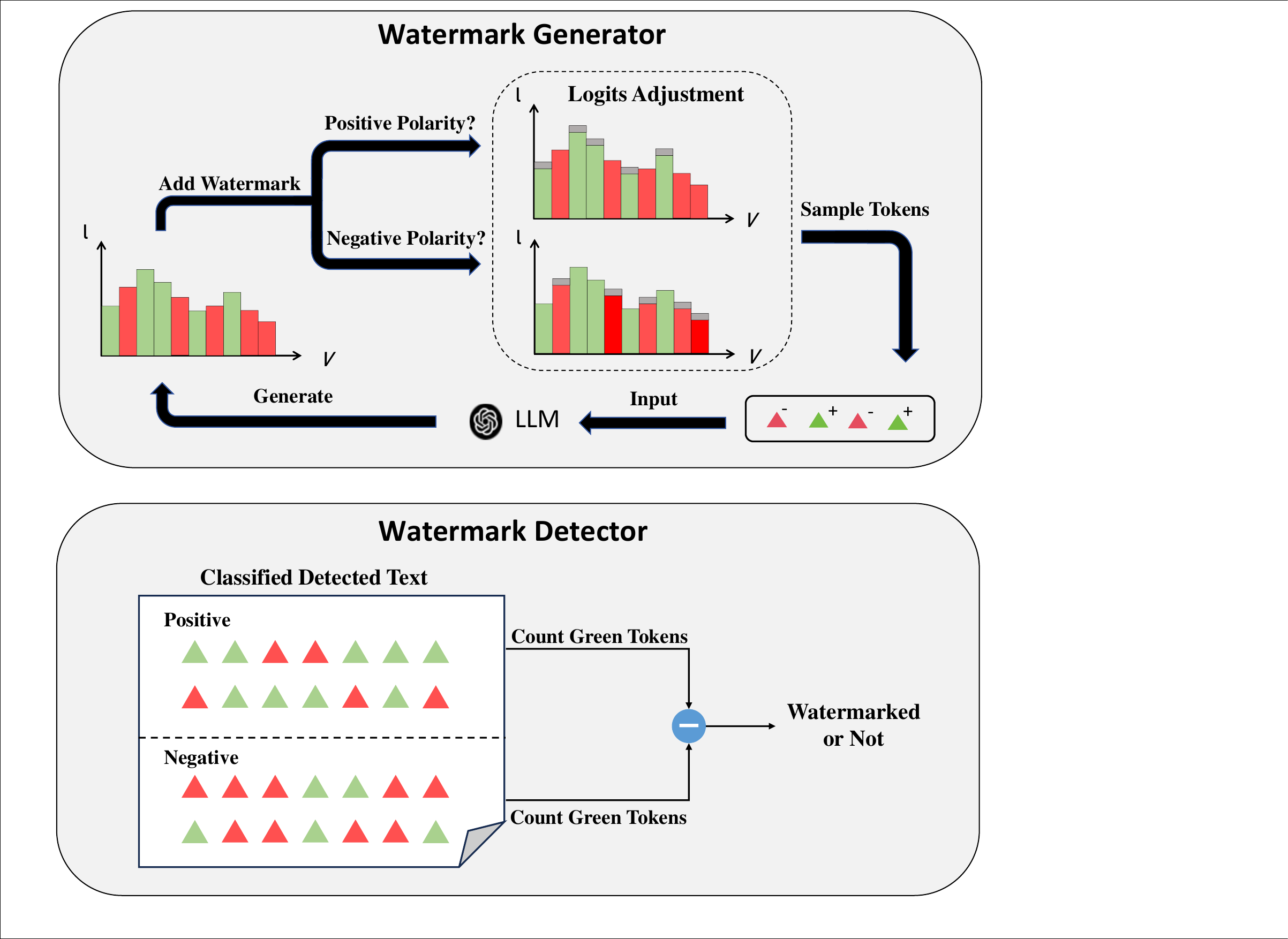}
\caption{Illustration of the key idea of \tool: Adjust token bias by polarity during generation and detect by the green token difference between polarities.}
\label{fig:overall}
\end{center}
\vskip -0.2in
\end{figure}

Watermarking techniques offer a solution by enabling LLMs to embed unique, imperceptible identifiers (watermarks) within generated content, distinguishing it from human-written text~\cite{liu2024survey,pan2024waterseeker}. A prominent method, \kgw\footnote{Details can be found in Appendix~\ref{app:pre-kgw}}, achieves high detection performance with low false positive and false negative rates~\cite{KGW}. \kgw\ works by partitioning the model's vocabulary into two categories: \textit{green} and \textit{red} tokens, with proportions \( \gamma \) and \( 1 - \gamma \), respectively. The model is then biased towards green tokens by adjusting their logits with a positive constant (watermark strength), ensuring that the generated text predominantly contains green tokens.
During detection, \kgw\ assumes that the number of green tokens in non-watermarked text follows a Gaussian distribution \( \mathcal{N}(\mu, \sigma^2) \), where \( \mu = T \cdot \gamma \) and \( \sigma^2 = T \cdot \gamma \cdot (1 - \gamma) \), with \( T \) representing the number of tokens analyzed. The likelihood that the text is watermarked increases as the observed number of green tokens deviates from the expected \( T \cdot \gamma \). In other words, the greater the number of green tokens in the analyzed text exceeds the expected value of \( T \cdot \gamma \) for non-watermarked text, the higher the probability that the text is watermarked.

Although \kgw\ demonstrated high detection performance, it exhibited weaknesses under more rigorous detection conditions, particularly in low-entropy scenarios~\cite{sweet}. Furthermore, it is crucial for detection methods to effectively distinguish between human-written and watermarked text to achieve higher accuracy while minimizing false positives, as false positives are intolerable in many contexts~\cite{bricks,sweet,KGW-RELIABLE}, such as falsely accusing a user of producing fake news or a student of cheating in an exam. A straightforward strategy is to increase watermark strength, which can help address this issue by enhancing the distinguishability between watermarked and non-watermarked texts. However, this improvement often comes at the cost of text quality~\cite{waterbench}.

Recent studies have proposed algorithms aimed at enhancing the statistical significance of detected watermarked text by leveraging its relationship with entropy~\cite{sweet,EWD}. For example, \citet{sweet} introduced a selective watermark detector called \sweet, specifically designed for code generation tasks, to improve watermark detectability in code. This method detects only those tokens that exceed a certain entropy threshold, thereby excluding tokens less likely to be influenced by the watermark generator. However, these algorithms depend on an auxiliary language model for vocabulary partitioning, which proves impractical when the tokenizers of the primary and auxiliary models are inconsistent~\cite{yoo-etal-2024-advancing}. Moreover, retrieving entropy values during detection necessitates access to the prompt's content. While the effectiveness of custom prompts has been demonstrated in code-related tasks~\cite{sweet,EWD}, applying this approach to general tasks poses significant challenges.

Orthogonal to prior work, this paper proposes a novel perspective for enhancing the distinction between watermarked and non-watermarked texts. Our approach is based on the analysis that existing methods often overlook the impact of non-watermarked text distributions across different scenarios on detection outcomes (see Section~\ref{sec:movitation}).
Specifically, taking \kgw\ as an example, the held assumption that the number of green tokens in non-watermarked text has an expected value of \( T \cdot \gamma \) does not consistently hold across various contexts. 
In some cases, this assumption undermines detection effectiveness\footnote{Appendix~\ref{app:example} provides an example illustrating the detection challenges arising from inaccurate estimations of non-watermarked text.}. For example, in low-entropy scenarios, both watermarked and non-watermarked texts tend to have a similarly small expected number of green tokens. 
Continuing to assume this fixed value as the expected value for non-watermarked text results in a lack of statistical significance, thereby increasing the risk of failing to effectively distinguish between watermarked and non-watermarked text.

To address the challenges posed by inaccurate green token count estimations in non-watermarked text that affect detection results, we propose \textit{Bi}polar Water\textit{Marker} (abbreviated as \tool). As illustrated in Figure~\ref{fig:overall}, the generated text is divided into two components: the \textit{positive pole} and the \textit{negative pole}. In the positive pole, the logits of tokens in the green list are increased by adding a positive constant, while in the negative pole, the logits of tokens in the red list are adjusted similarly. 
For detection, instead of comparing the number of green tokens to the ideal expected count, we calculate the green token counts in both poles and determine their difference. This approach significantly enhances the distinction between watermarked and human-written text. Moreover, through various analyses, we demonstrate that our method is orthogonal to existing techniques that leverage entropy or enhance watermark detection. It can be effectively combined with these optimization techniques in scenarios where entropy can be computed and prompts are available.

In summary, the key contributions of our work are summarized as follows: 
\begin{itemize} 

\item New Dimension: We reveal the impact of inaccurate estimations of human-written text distributions on detection results, offering a new perspective for enhancing the detectability of watermarks.

\item New Method: We introduce \tool, a novel bipolar watermarking method with differential detection. This approach effectively addresses the challenges posed by the estimation inaccuracies of human-written text distributions, enhancing the distinction between watermarked and human-generated content.

\item Theoretical Proof: We provide a comprehensive theoretical analysis of detection accuracy, demonstrating that our method improves the detectability of watermarked text without increasing the false positive rate, and that it is equally effective when used in conjunction with other optimization techniques.

\item Experimental Validation: Our experiments demonstrate that \tool\ surpasses traditional watermarking methods in detection accuracy. Additionally, it is fully compatible with entropy-based watermark optimization techniques.
\end{itemize}


\section{Related Works}

Text watermarking is a form of linguistic steganography, where the objective is to embed a hidden message—the watermark—within a piece of text. Typically, text watermarking is divided into two main categories: watermarking for existing texts and watermarking for Large Language Models (LLMs).

\subsection{Watermarking for Existing Texts}

Some methods embed watermarks by modifying the text format rather than its content, such as line or word-shift coding or Unicode modifications like whitespace replacement (WhiteMark) and variation selectors (VariantMark)\cite{format-1,format-2}. While these approaches preserve the meaning of the text and allow for high payloads, they are vulnerable to removal through canonicalization (e.g., resetting spacing)\cite{format-3} and can be easily forged, which compromises their reliability~\cite{format-4}.

Some watermarking methods modify the content of the text, such as by changing words~\cite{lexical-1,lexical-2,robustness,lexical-4,lexical-5}. Yang et al.\cite{lexical-4} introduced a BERT-based in-fill model to generate contextually appropriate lexical substitutions, along with a watermark detection algorithm that identifies watermark-bearing words and applies inverse rules to extract the hidden message. In another work, Yang et al.\cite{lexical-5} simplified detection by encoding words with random binary values, substituting bit-0 words with context-based synonyms for bit-1. However, these lexical substitution methods can be vulnerable to simple removal techniques, such as random synonym replacement.

\subsection{Watermarking for LLMs}
Watermark embedding during LLM generation can be achieved by modifying token sampling~\cite{sample-2,sample-1} or model logits~\cite{KGW,yoo-etal-2024-advancing,robustness-3,EWD}, resulting in more natural text~\cite{markllm}. \kgw~\cite{KGW} represents a promising approach for distinguishing language model outputs from human text while maintaining robustness against realistic attacks, as it can reinforce the watermark with every token. Several studies have built upon \kgw, addressing various aspects such as low-entropy generation tasks like code generation~\cite{sweet,EWD}, enhancing watermark undetectability~\cite{Christ_2023}, enhancing performance in payload~\cite{yoo-etal-2024-advancing,codable} and improving robustness~\cite{robustness,robustness-1,robustness-2,robustness-3}. For example, ~\citet{bricks} expanded from a binary red-green partition to a multi-color partition through fine-grained vocabulary partitioning, allowing for the embedding of multi-bit watermarks. Additionally, ~\citet{robustness-3} converted semantic embeddings into semantic values using weighted embedding pooling, followed by discretization with NE-Ring, and classified the vocabulary into red-list and green-list categories based on these semantic values, thus enhancing robustness.

Some existing methods aim to improve the detectability of \kgw\ in low-entropy scenarios~\cite{EWD,sweet}.
For instance, \citet{EWD} addresses this by assigning weights to tokens based on their entropy during detection, thereby enhancing sensitivity by emphasizing high-entropy tokens in z-score calculations. In contrast, our approach does not require using large language models to compute token entropy during detection, nor does it depend on the availability of prompt content. Moreover, our method is orthogonal to these approaches, allowing it to be seamlessly integrated with such techniques for enhanced performance. 

\section{Method}

\begin{figure}[ht]
\vskip 0.2in
\begin{center}
\centering\includegraphics[scale=0.45,trim= 0cm 0cm 1cm 1.4cm, clip]{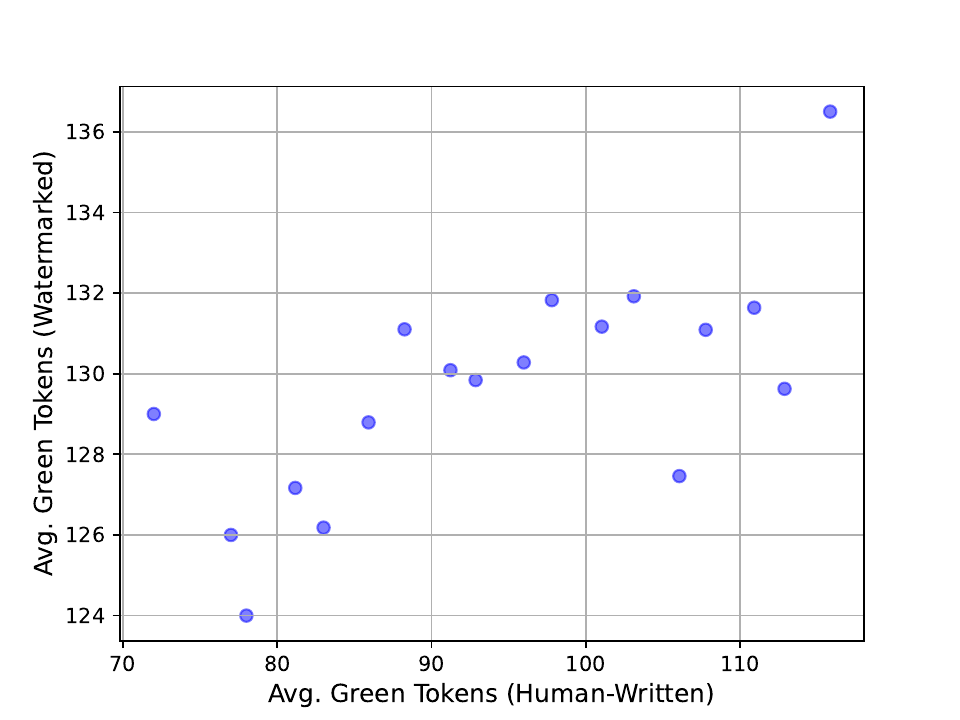}
\caption{Empirical Study of the Relationship Between the Number of Green Tokens in Human-Written Texts and the Number of Green Tokens in Watermarked Texts.}
\label{movitation}
\end{center}
\vskip -0.2in
\end{figure}

\subsection{Movitation}
\label{sec:movitation}

Before introducing our method, we first discuss the distribution of human-written text. It is undeniable that the distribution of green tokens in human-written text should be \( \gamma T \), but in more fine-grained scenarios, this may not always be accurate. 
Furthermore, we present our assumption: \textit{the expected number of green tokens in human-written text and watermarked text follow a similar trend}. In some scenarios, if the expected number of green tokens in the watermarked text is lower, the corresponding human-written text is likely to exhibit a similarly lower expected number of green tokens.

To validate that the number of green tokens in human-written and watermarked texts follows a similar pattern, we conducted a simple empirical study. We used the \kgw\ watermarking algorithm (\( \gamma = 0.5 \), \( \delta = 1 \), and polynomial sampling) with OPT-1.3B~\cite{opt} to generate 500 watermarked text samples, each consisting of 200 tokens. The prompt content was taken from the RealNewsLike subset of the C4 dataset~\cite{c4}. For each sample from this dataset, we recorded the last 200 tokens, which represent the human-written text under the same prompt as the watermarked text. The remaining tokens served as the prompt.

We selected a subset of the experimental results where the number of green tokens in human-written texts was fewer than 120, grouping them in intervals of 2.5 based on this count. For each group, we computed the average number of green tokens in both the human-written and watermarked texts. The results, shown in Figure~\ref{movitation}, demonstrate that when the number of green tokens in human-written text is low, the watermarked text exhibits a similar trend, with fewer green tokens. A Pearson correlation coefficient of 0.7 indicates a strong positive correlation between the two distributions.



However, \kgw\ considers only the theoretical lower bound of green tokens in watermarked text under different entropy conditions, overlooking the fact that the expected number of green tokens in human-written text also varies. During detection, it evaluates the difference between the number of green tokens in the analyzed text and the fixed expected value \( \gamma T \), with larger deviations indicating stronger statistical significance. However, treating the number of green tokens in human-written text as a fixed reference value \( \gamma T \) is overly simplistic and imprecise.

To overcome the impact of inaccurate estimates of green token distributions in human-written text on detection, 
we introduce~\tool. Since prompt content and corresponding human-written text are often unavailable during detection, directly determining the expected number of green tokens for a given prompt is not feasible. 
Instead, we divide the text under detection into two components—the positive and negative poles (i.e., each token in the text belongs to either the positive or negative pole)—and detect the watermark based on their differences. 
For instance, when \( \gamma = 0.5 \), we divide the text into two parts with an equal number of tokens. Under ideal partitioning, the expected difference in the number of green tokens between the positive and negative poles in human-written text is zero. 
This approach incurs no additional inference cost, does not depend on prompt content, and maintains the same time complexity, making it applicable across various scenarios. In the following sections, we detail our method and explain its effectiveness.



\subsection{Our Method}
\paragraph{Generating the Watermark.} Our watermark embedding algorithm, as outlined in Algorithm~\ref{alg:generator}, 
differs from the \kgw\ algorithm, which uniformly boosts the scores of green tokens across all generated text. Instead, our approach divides the generated text into positive and negative polarities. In the positive polarity phase, the algorithm increases the logits of tokens in the green list to enhance their sampling probabilities.
Conversely, in the negative polarity phase, it increases the logits of tokens in the red list, thereby reducing the sampling probabilities of green tokens. The polarity (positive or negative) is pseudo-randomly determined based on the probability \(\rho\), where \(\rho\) represents the probability of the positive polarity. 
This approach avoids hardcoding the polarities into the input text sequence, thereby reducing the risk of detection failure caused by confusion in non-watermarked text scenarios.
Notably, our method consistently increases logits across $\gamma$ proportion of the vocabulary, resulting in a distribution change for the generated text that is consistent with \kgw. 
Notably, our method consistently increases logits across a \(\gamma\) proportion of the vocabulary, resulting in an overall distribution change for the generated text that aligns with \kgw.
This implies that under identical \(\gamma\) and \(\delta\), our method maintains the same variations in text quality as \kgw (see Appendix~\ref{ppl}).
\paragraph{Detecting the watermark.} 
\label{sec:detect}
We can detect the watermark by testing the following null hypothesis 
$H_0$: \textit{The text sequence is generated without any knowledge of the red list rule or the polarity rule.}


\begin{algorithm}[tb]
   \caption{Text Generation with Bipolar Watermarks}
   \label{alg:generator}
\begin{algorithmic}
   \STATE {\bfseries Input:} prompt $x_{prompt}$, $\gamma \in (0, 1)$, $\delta>0$,  $\rho \in (0, 1)$
   \FOR{$i=0,1,2,... $}
   \STATE Compute a logit vector $l_i$ by (\ref{eq:logits});
   \STATE Compute a hash of token $s_{:i}$, and use it to seed a random number generator;
   \STATE Using this random number generator, randomly partition the vocabulary into two lists, $list1$ and $list2$, with sizes $\gamma|V|$ and $(1 - \gamma)|V|$, respectively;
   \STATE Use the generator to decide the current polarity $p$ (positive or negative) according to the probability $\rho$;
   \IF{$p$ is positive}
  \STATE $list1$ is the green token list $G$, and $list2$ is the red token list $R$;
   \STATE Add $\delta$ to the logits of tokens in $G$;
   \ELSE
     \STATE $list2$ is the green token list $G$, and $list1$ is the red token list $R$;
    \STATE Add $\delta$ to the logits of tokens in $R$;
   \ENDIF
   \STATE Sample $s_l$ by (\ref{eq:sample});
   \ENDFOR
\end{algorithmic}
\end{algorithm}

During detection, we count the number of positive and negative green tokens, denoted as \( |s|_{pG} \) and \( |s|_{nG} \), respectively. Let \( T_p \) and \( T_n \) represent the total numbers of positive and negative tokens. Under the null hypothesis (\( H_0 \)), when the distributions of the positive and negative poles are independent, the expected value of \( |s|_{pG} - |s|_{nG} \) is \( T_p \cdot \gamma - T_n \cdot (1 - \gamma) \), with a variance of \( T \cdot \gamma \cdot (1 - \gamma) \), where \( T = T_p + T_n \). The z-statistic for this test is calculated as:  
\begin{equation}
z = \frac{|s|_{pG} - |s|_{nG}-\gamma T_p+(1-\gamma) T_n}{\sqrt{T\gamma(1-\gamma)}}.
\label{eq:z_d}
\end{equation}

\paragraph{Remark.}
From Formula~\ref{eq:z_d}, we observe that our numerator still relies on the estimated expectation for the number of green tokens in non-watermarked text. However, this influence is mitigated due to the subtraction of the estimated values for positive and negative polarities. When the number of text sequences generated for the positive and negative polarities satisfies the ratio \( T_p / T_n = (1 - \gamma) / \gamma \), the \( z \)-value simplifies to:  

\[
z = \frac{|s|_{pG} - |s|_{nG}}{\sqrt{T \cdot \gamma (1-\gamma)}}.
\]

In this scenario, the \( z \)-value directly reflects the observed difference between the counts of positive and negative green tokens. This reduces the reliance on estimated values for non-watermarked text, thereby diminishing the impact of potential inaccuracies in these estimations on the final detection result.


\subsection{Effect of Bipolar Watermark}

This section demonstrates that our \tool\ improves detectability. Theorem~\ref{theorem:low-bound-green} indicates that a higher lower bound of the \( z \)-score can be achieved using \tool\ compared to \kgw. This improvement is realized by reducing the decline in detection accuracy caused by inaccurate estimations of non-watermarked text.

\begin{theorem}
\label{theorem:low-bound-green}
Consider watermarked text sequences \( s \) of \( T \) tokens. Each sequence is produced by sequentially sampling a raw probability vector \( p(t) \) from the language model, sampling a random green list of size \( \gamma |V| \). The sequence \( s \) is composed of a positive part \( s_p \) and a negative part \( s_n \).
The logits for the green and red lists are boosted by \( \delta \) before sampling each token in the positive and negative parts, respectively. Define \( \alpha = \exp(\delta) \), and let \( |s|_d \) denote the difference in the number of green list tokens in the positive and negative parts of sequence \( s \).
We assume that positive and negative tokens are mutually independent and identically distributed, then \tool\ achieves a theoretical lower bound for the \( z \)-score that consistently exceeds that of \kgw.
\end{theorem}

Furthermore, we present the following theorem regarding non-watermarked text: our differential detection method does not increase the false positive rate.

\begin{theorem} 
\label{theorem:no-increase-false-positive} 

We assume that the total number of green tokens in non-watermarked text follows a Gaussian distribution \( \mathcal{N}(\mu_T, \sigma_T^2) \), while the counts of positive and negative tokens follow \( \mathcal{N}(\mu_{T_p}, \sigma_{T_p}^2) \) and \( \mathcal{N}(\mu_{T_n}, \sigma_{T_n}^2) \), respectively, with the distributions being independent. Additionally, we have \( T_p / T_n = (1 - \gamma) / \gamma \).

We denote the false positive rates of the unipolar and differential detection methods as \( F_{KGW} \) and \( F_{DIFF} \), respectively, leading to the relationship:

\[
F_{KGW} \geq F_{DIFF}.
\]

\end{theorem}

The detailed proofs of Theorems~\ref{theorem:low-bound-green} and \ref{theorem:no-increase-false-positive} are provided in Appendix~\ref{app:proof}.

\paragraph{Applying \tool\ to \sweet\ and \ewd.}


\sweet\ and \ewd\ are advanced algorithms derived from \kgw, designed to enhance the statistical significance of watermark text by leveraging entropy. Our proposed algorithm is orthogonal to both of these approaches. The application of our bipolar watermarking and differential detection algorithms, based on \sweet\ and \ewd, is detailed in Appendix~\ref{app:to-sweet-ewd}. Moreover, the conclusions of Theorem~\ref{theorem:low-bound-green} and Theorem~\ref{theorem:no-increase-false-positive} remain valid when our algorithms are applied to \sweet\ and \ewd. In other words, our algorithm enhances the statistical significance of watermark text in both \sweet\ and \ewd\ while preserving the original false positive rate.
Below, we elaborate on the effectiveness of applying the bipolar watermarking and differential detection algorithms to \sweet, while the rationale for their application to \ewd\ is explained in Appendix~\ref{app:proof}.

The primary distinction between \sweet\ and \kgw\ lies in the fact that \sweet\ applies a bias \( \delta \) to the logits of tokens in \( G \) only if their entropy exceeds a specified threshold \( \tau \). This mechanism enables the algorithm to selectively target subsequences of the generated text based on their entropy levels. According to Theorem~\ref{theorem:low-bound-green} and Theorem~\ref{theorem:no-increase-false-positive}, when such subsequences satisfy the conditions outlined in our theoretical framework, the following conclusions can be drawn: applying \sweet\ with our bipolar watermarking and differential detection algorithms results in higher theoretical lower bounds for the \( z \)-values, without increasing the false positive rate.

\section{Experiments}

In this section, we first evaluate the performance of \tool\ and \kgw\ under high-entropy conditions to demonstrate the effectiveness of our watermarking algorithm. Subsequently, we assess the performance of \tool\ when used in conjunction with existing \kgw-based optimization algorithms, \sweet\ and \ewd, in code-related tasks, which are their primary focus. This evaluation illustrates that our algorithm is orthogonal to other optimization methods, providing new insights for future optimization approaches.

To assess the performance of the watermark, we primarily analyze the True Positive Rate (TPR) and False Positive Rate (FPR), where TPR represents the detection of watermarked text, and FPR indicates human text falsely flagged as watermarked. We also present the F1 score as an additional metric. 
Furthermore, we provide empirical evidence in Appendix~\ref{app:curve} showing that our watermarking method does not degrade the quality of the generated text. For non-code tasks, we follow \citet{KGW} and use use PPL as the evaluation metric. For code generation tasks, we adopt pass@1 accuracy to measure functional correctness, following \citet{sweet}.
Our implementation is based on the completion of \kgw\ with the probability ratio of positive and negative polarities set to \((1 - \gamma) / \gamma\).

\paragraph{Tasks and Datasets.}
Our experiments are designed with two scenarios. The first scenario is a high-entropy environment, where we use the same experimental dataset as \kgw\ and employ the OPT-1.3B~\cite{opt} model to generate watermarked texts. To simulate diverse and realistic language modeling conditions, we randomly select texts from the news-like subset of the C4 dataset~\cite{c4}. Following \kgw's setup, for each randomly selected string, a fixed number of tokens are trimmed from the end to serve as the baseline completion, with the remaining tokens acting as the prompt. 
We then utilize a larger oracle language model (OPT-2.7B) to compute the perplexity (PPL) of the generated completions. 
In all experiments, we generate 500 samples, each with a length of approximately 200 ± 5 tokens.

The second scenario focuses on low-entropy conditions, which are the primary target of \sweet\ and \ewd. For this, we adopt the task of code generation, using two datasets: HumanEval~\cite{HumanEval} and MBPP~\cite{mbpp}, 
following ~\citet{sweet}. 
These datasets consist of Python programming problems 
with test cases and corresponding reference answers, 
which are treated as human-written samples. 
Code generation is conducted using StarCoder~\cite{starcoder}. 
During evaluation, the length of both the generated 
and reference samples is restricted to at least 15 tokens.
For more detailed experimental settings, please refer to Appendix~\ref{app:experiment-setting}.


\begin{figure*}[ht]
\vskip 0.2in
\begin{center}
\centerline{\includegraphics[width=1.5\columnwidth]{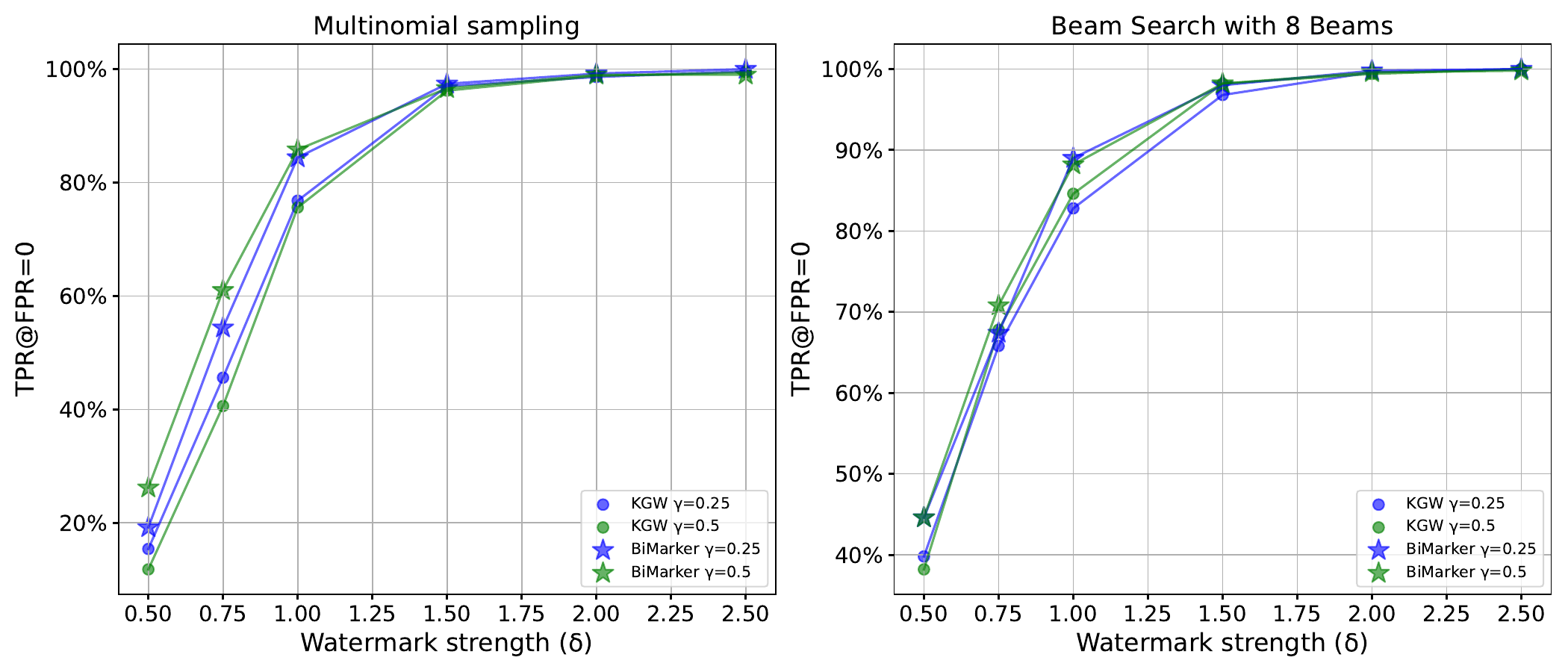}}
\caption{The relationship between watermark strength and accuracy under zero false positives (0/500). The left figure illustrates results with multinomial sampling, while the right figure depicts results using beam search with a beam size of 8.}
\label{ppl-tp}
\end{center}
\vskip -0.2in
\end{figure*}

\begin{table}[t]
\caption{TPR at various FPR for watermark detection using \kgw\ and \tool\ are presented, with \(\gamma = 0.5\). The first three comparisons employ multinomial sampling, while the final comparison utilizes beam search with a beam size of 8.}
\label{tp-table}
\vskip 0.15in
\begin{center}
\begin{small}
\begin{sc}
\begin{tabular}{lccccc}
\toprule
   &  &\multicolumn{2}{c}{$1\%$FPR } & \multicolumn{2}{c}{$5\%$FPR } \\  \cline{3-6}
 \raisebox{1.5ex}{Methods} &\raisebox{1.5ex}{$\delta$} &TPR  &F1 &TPR &F1 \\ 
\midrule
  \kgw\    & 0.5  &0.436&0.603 &0.7 &0.8\\
\tool\ & 0.5  &\textbf{0.498}&\textbf{0.66}  &\textbf{0.744} &\textbf{0.829}\\ \cline{1-6}

  \kgw\    & 1.5  &0.978          &0.984         &0.992  &0.972\\
  \tool\ & 1.5  &\textbf{0.986} &\textbf{0.988}&\textbf{0.992} &\textbf{0.972}\\ \cline{1-6}

  \kgw\    & 2.0  &0.99          &0.99           &\textbf{0.994} &0.973\\
\tool\ & 2.0  &\textbf{0.994}&\textbf{0.992} &\textbf{0.994} &\textbf{0.973}\\ \cline{1-6}
  \kgw\    & 2.0  &0.998            &0.994           &1.0 &0.976\\
\tool\ & 2.0  &\textbf{0.998}   &\textbf{0.994}  &\textbf{1.0} &\textbf{0.976}\\


\bottomrule
\end{tabular}
\end{sc}
\end{small}
\end{center}
\vskip -0.1in
\end{table}

\paragraph{Main Results.} Figure~\ref{ppl-tp} illustrates the relationship between watermark strength (represented by $\delta$) and accuracy under various hyperparameter settings. Additional metrics, including ROC curves, are provided in Appendix~\ref{app:curve}.
We consider combinations of $\gamma$ = [0.25, 0.5] and $\delta$ = [0.25, 0.75, 1, 1.5, 2, 2.5], utilizing both multinomial sampling and 8-beam search. It is clear that larger $\delta$ values result in stronger watermarking, but at the cost of text quality. 
Given that incorrectly labeling human-written texts as watermarked carries more significant consequences than falsely identifying watermarked texts as human-written~\cite{bricks,sweet,KGW-RELIABLE}, we focus on the accuracy in scenarios where no false positives (0/500) are observed.


From the figure, we observe that \tool\ outperforms \kgw\ in distinguishing watermarked texts from human-written texts, particularly when the watermarking strength is low. The enhancement in performance is more pronounced under multinomial sampling compared to 8-beam search. 
Notably, under multinomial sampling, when \(\delta\) is set to 1, the TPR increases by 10\% with no false positives, and at \(\delta = 0.75\), it can even increase by 20\%.
In contrast, when $\delta$ is set to 2.5 under 8-beam search, both methods achieve a TPR of 1.
Table~\ref{tp-table} presents the accuracy and F1 scores under TPR at 1\% and 5\% false positive rates for different watermark strengths when $\gamma = 0.5$. The observed trends are consistent with those in Figure~\ref{ppl-tp}. 
These results demonstrate the effectiveness of \tool\ across various hyperparameter settings.

\begin{figure}[ht]
\vskip 0.2in
\begin{center}
\centerline{\includegraphics[scale=0.5,trim= 0.4cm 0.1cm 1.5cm 1.4cm, clip]{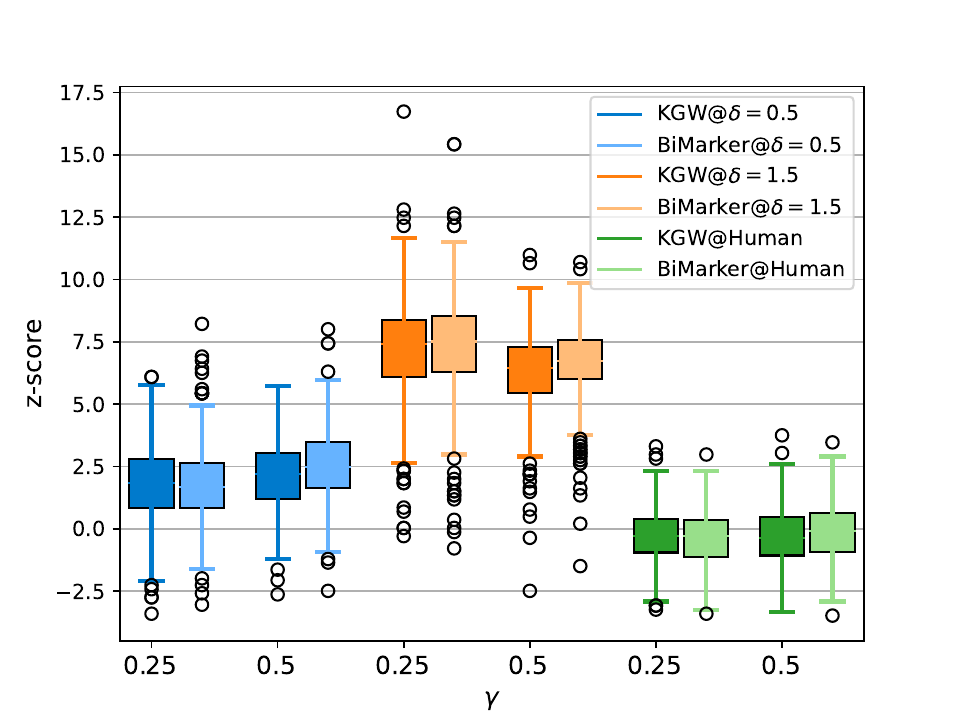}}
\caption{Z-scores of Watermarked and Human Texts under Different Parameter Settings with multinomial sampling.}
\label{fig:z_score}
\end{center}
\vskip -0.2in
\end{figure}

\paragraph{Analysis of the Reasons for Performance Improvement.}
Figure~\ref{fig:z_score} shows the z-scores of both watermarked and human texts under different hyperparameter settings, detected by various methods. Our \tool\ method results in overall higher z-scores for watermarked texts, while the z-scores for human texts remain relatively unchanged. By enlarging the gap between the z-scores of watermarked and human texts, our \tool\ demonstrates an enhanced ability to distinguish watermarked texts effectively.
Furthermore, these results support the validity of our Theorem~\ref{theorem:low-bound-green} and Theorem~\ref{theorem:no-increase-false-positive}.

\begin{table*}[t]
\caption{Comparison of Detection Results Before and After Applying the \tool\ Algorithm to \kgw, \sweet\ and \ewd\ Across Different Datasets ($\gamma=0.5, \delta=2$)}
\label{tab:low-entroy}
\vskip 0.15in
\begin{center}
\begin{small}
\begin{sc}
\begin{tabular}{lcccccccccc}
\toprule
         & \multicolumn{5}{c}{HumanEval}&\multicolumn{5}{c}{MBPP} \\  \cline{2-11}
Methods  & \multicolumn{2}{c}{1\%FPR}   &\multicolumn{2}{c}{5\%FPR}&Best  & \multicolumn{2}{c}{1\%FPR} &\multicolumn{2}{c}{5\%FPR}&Best  \\ \cline{2-11}
         & TPR &F1 &TPR &F1 &F1 &TPR &F1 &TPR &F1 &F1 \\ 
\midrule
  \kgw\          &\textbf{0.375}    &\textbf{0.542} &0.5             &0.645          &0.787      & 0.098         &0.177         & 0.347    &0.497&0.739        \\
\kgw+\tool\      &0.352      &0.516 &\textbf{0.609}  &\textbf{0.734} &\textbf{0.797} &\textbf{0.176} &\textbf{0.296}& \textbf{0.38}    &\textbf{0.531}&\textbf{0.749}        \\ \cline{1-11}
 \sweet\            &0.561         &0.714 &0.758 &0.838&\textbf{0.875} & 0.381               &0.548    &\textbf{0.686} &\textbf{0.791}&0.85        \\
\sweet+\tool\    &\textbf{0.563}&\textbf{0.715}  &\textbf{0.758} &\textbf{0.838} &0.872   &\textbf{0.415}                & \textbf{0.582}        &0.67      &0.779     &\textbf{0.852}   \\ \cline{1-11}
\ewd\             &0.609    &0.753 &0.75  &0.833&0.869 & 0.6  &0.746    &0.762    &0.841 &0.888 \\
\ewd+\tool\      &\textbf{0.671}&\textbf{0.799} &\textbf{0.75}  &\textbf{0.833}&\textbf{0.96}  & \textbf{0.618}&\textbf{0.759}    &\textbf{0.833}  &\textbf{0.844}  &\textbf{0.898} \\ 

\bottomrule
\end{tabular}
\end{sc}
\end{small}
\end{center}
\vskip -0.1in
\end{table*}

\begin{figure}[ht]
\vskip 0.2in
\begin{center}
\centering\includegraphics[scale=0.5,trim= 0.9cm 0.1cm 1.5cm 1.4cm, clip]{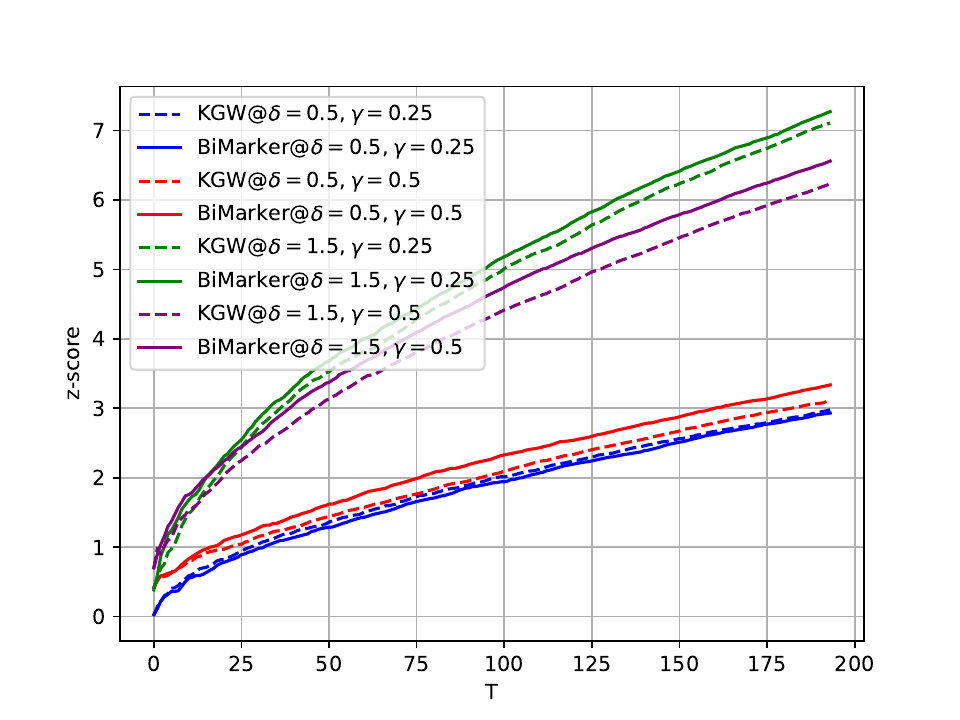}
\caption{The Average Z-Score as a Function of Token Length (\(T\)) of the Generated Text with multinomial sampling.}
\label{fig:loc-z}
\end{center}
\vskip -0.2in
\end{figure}

\paragraph{Performance with Different Numbers of Tokens.}
Figure~\ref{fig:loc-z} shows the detected average z-score over samples as \( T \) varies. The curves are presented for various values of \( \delta \) and \( \gamma \) using multinomial sampling. We observe that, compared to \kgw, \tool\ does not sacrifice its detection capability on short text sequences. \tool\ consistently achieves higher z-scores for watermarked texts across different values of \( T \) compared to \kgw.

\begin{table}[t]
\caption{Detection performance of back-translated watermarked texts using different watermarking algorithms, with $\gamma = 0.5$.}
\label{table:attack}
\vskip 0.15in
\begin{center}
\begin{small}
\begin{sc}
\begin{tabular}{lccccc}
\toprule
\multicolumn{6}{c}{Detection w/ Back-translation} \\ \cline{1-6}
   &   &\multicolumn{2}{c}{$1\%$FPR } & \multicolumn{2}{c}{$5\%$FPR }  \\  \cline{3-6}
 \raisebox{1.5ex}{Methods} & \raisebox{1.5ex}{$\delta$} &TPR  &F1 &TPR &F1\\ 
\midrule
  \kgw\   & 0.5  &0.188         &0.314              &0.488  &         0.635\\
  \tool\  & 0.5  &\textbf{0.272}&\textbf{0.424}     &\textbf{0.496}  &\textbf{0.642}    \\ \cline{1-6}
  \kgw\   &1.5   &\textbf{0.786}& \textbf{0.875}        &0.914              &0.931     \\
  \tool\ &1.5  &  0.776        & 0.868                 &\textbf{0.93}      &\textbf{0.939}     \\ 
\bottomrule
\end{tabular}
\end{sc}
\end{small}
\end{center}
\vskip -0.1in
\end{table}

\paragraph{Performance against the Back-translation Attack.}
Watermarked texts are often edited before detection, which can partially remove the watermark and degrade detection performance. To evaluate the robustness of our \tool\ under these conditions, we simulate this scenario by employing back-translation as an attack strategy to alter the watermarked text. Specifically, we first generate watermarked texts and then translate them from English to French before translating them back to English for detection. This approach not only mirrors realistic adversarial scenarios but also ensures that the modifications aim to obscure the watermark while preserving the semantic integrity of the original text. 
Table~\ref{table:attack} presents our detection results with back-translation. Our experiments show that \tool\ achieves overall better detection accuracy and demonstrates performance comparable to \kgw.

\begin{figure}[ht]
\vskip 0.2in
\begin{center}
\centerline{\includegraphics[scale=0.5,trim= 0.4cm 0.1cm 1.5cm 1.4cm, clip]{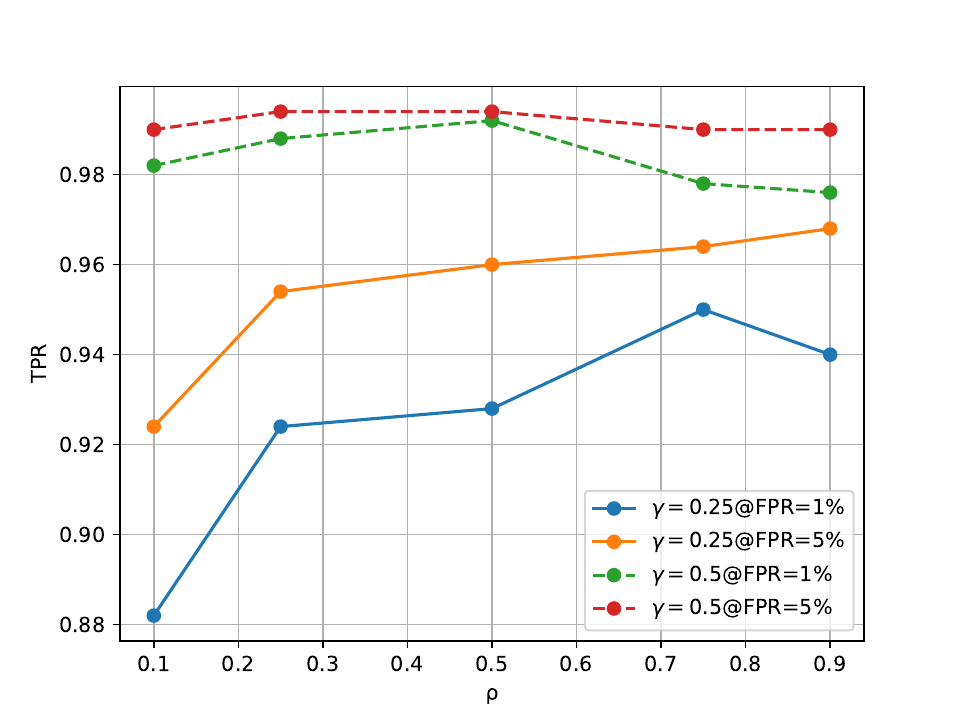}}
\caption{The impact of $\rho$ values on detection results. We set \(\delta = 1.5\) and used multinomial sampling.}
\label{fig:rou}
\end{center}
\vskip -0.2in
\end{figure}

\paragraph{Impact of \(\rho\) on Results.}

In Section~\ref{sec:detect}, we established that watermark detection performance is maximized when \( T_p / T_n = (1 - \gamma) / \gamma \). To verify this conclusion, we conducted experiments. By setting \(\gamma = 0.25\) and \(0.5\), the theoretically optimal \(\rho\), representing the positive ratio, is \(0.75\) and \(0.5\), respectively. We varied \(\rho\) values to compare detection accuracy. To minimize the impact of randomness, we adopted a position-based hard-coded division: the first \(200 \cdot \rho\) tokens were assigned to the positive polarity, while the remaining tokens were assigned to the negative polarity. The experimental results, shown in Figure~\ref{fig:rou}, confirm our theoretical prediction, validating the correctness of our theory.


\paragraph{Application of \tool\ to \sweet\ and \ewd.}
Embedding watermarks in low-entropy texts presents a significant challenge. We evaluate whether our bipolar watermarking and differential detection methods can be applied to existing algorithms, including \kgw\ and its variants, \sweet\ and \ewd, with the goal of enhancing their performance and paving the way for future research directions. Specifically, when applied to \kgw\ and \ewd, we employ position-based hard encoding for the polarities, using a repeating cycle of 20 positive tokens followed by 20 negative tokens. However, for \sweet, which applies watermark embedding and detection only to high-entropy segments, we select a smaller cycle of 15 positive tokens followed by 15 negative tokens.
Table~\ref{tab:low-entroy} presents the results before and after the application of  \tool, showing a significant improvement in detection effectiveness. For instance, when applied to the current state-of-the-art algorithm \ewd, our \tool\ demonstrates enhancements across all metrics for each dataset.

\section{Conclusion}
In this work, we reveal the impact of the inaccurate estimation of the number of green tokens in non-watermarked texts by existing watermarking algorithms on watermark detection accuracy. To address this issue, we propose a bipolar watermarking and differential detection algorithm called \tool. Instead of comparing the number of green tokens to a fixed value, \tool\ calculates the green token counts in both poles and determines their difference. We conduct both theoretical and experimental analyses to evaluate the effectiveness of \tool. The results demonstrate that this algorithm effectively enhances the detectability of watermarks, exhibiting excellent performance, particularly under low watermark strength and low false positive rate conditions. Furthermore, we provide theoretical and experimental evidence showing that our algorithm is orthogonal to \sweet\ and \ewd\, indicating that its application can significantly improve the detectability of low-entropy texts when integrated with these algorithms. This opens up new avenues for future optimization efforts.

\section*{Limitations}


\tool\ uses hard-coded polarity assignment with \sweet\ and \ewd\ in low-entropy scenarios, assuming independent token distributions. However, code syntax introduces adjacent token dependencies—we mitigate this by enforcing minimum spacing between opposite polarities, though watermark integrity risks persist. Varying token counts also hinder optimal polarity ratios. Future work will refine polarity assignment robustness.

\nocite{langley00}

\bibliography{example_paper}
\bibliographystyle{icml2025}

\newpage
\appendix
\onecolumn

\section{Preliminaries}

\subsection{Language model basics}
Here, we introduce the fundamental concept of LLMs as well as the principles of embedding and detecting watermarks during the logits generation phase. An LLM, denoted as \( M \), takes a prompt as input and generates subsequent tokens as its response. Specifically, let the initial prompt be \( x_{\text{prompt}} \). At the \( i \)-th step, the input to the LLM consists of \( x_{\text{prompt}} \) and the sequence of already-generated tokens \( s_{:i-1} \). Based on this input, the LLM predicts the probability distribution for the next token \( t_i \) over the vocabulary \( V \), represented as logits \( l_i \): 

\begin{equation}
l_i = M(x_{\text{prompt}}, s_{:i-1}).
\label{eq:logits}
\end{equation}

The next token \( t_i \) is then selected from the predicted logits \( l_i \) using techniques such as nucleus sampling, greedy decoding, or beam search. The sampling process can be expressed as:

\begin{equation}
t_i = \mathcal{S}(\text{softmax}(l_i)),
\label{eq:sample}
\end{equation}

where \( \mathcal{S} \) denotes the specific sampling method used to choose the next token.

\subsection{Watermarking during Logits Generation}
\label{app:pre-kgw}
\kgw\ partitions the vocabulary into a red list (\( R \)) and a green list (\( G \)) at each token position, determined by a hash function that depends on the preceding token sequence. For the \( i \)-th token generation by the model, a bias \( \delta \) is applied to the logits of tokens in \( G \). The adjusted logit value \( \tilde{l}(i) \) for a token \( v \) at position \( i \) is calculated as follows:  

\begin{equation}
\tilde{l}(i)[v] = 
\begin{cases} 
l(i)[v] + \delta, & v \in G \\ 
l(i)[v], & v \in R 
\end{cases}
\label{eq:adjusted_logits}
\end{equation}

Here, \( l(i)[v] \) represents the original logit value for token \( v \) at position \( i \), and \( \tilde{l}(i)[v] \) is the modified logit value after applying the watermarking bias. 
To detect the watermark, the algorithm classifies each token as either green or red based on the same hash function used during watermark embedding. It then calculates the green token ratio and evaluates it using the z-metric, defined as:

\begin{equation}
z = \frac{\left|s\right|_G - T \cdot \gamma}{\sqrt{T \cdot \gamma \cdot (1 - \gamma)}},
\label{eq:z_score}
\end{equation}

Below, we discuss the concept of spike entropy, the theoretical lower bound for the number of green tokens embedded with watermarks, and the factors influencing perplexity.

\paragraph{Spike Entropy.} Spike Entropy as defined by \citet{KGW} is used for measuring how spread out a distribution is. Given a token probability vector $\mathbf{p}$ and a scalar $m$, the spike entropy of $\mathbf{p}$ with modulus $m$ is defined as:


\begin{equation}
S(\mathbf{p}, m) = \sum_k p_k \frac{1}{1 + m p_k}.
\label{eq:spikle}
\end{equation}

\paragraph{Theoretical Lower Bound for Green Tokens.}
We first introduce an important lemma from \citet{KGW}:
\begin{lemma}
\label{lemma:low-bound-green-kgw}
Suppose a language model produces a raw (pre-watermark) probability vector $p \in (0, 1)^N$. Randomly partition $\mathbf{p}$ into a green list of size $\gamma N$ and a red list of size $(1 - \gamma)N$. Form the corresponding watermarked distribution by boosting the green list logits by $\delta$. 
Define $\alpha = \exp(\delta)$.
The probability that the token is sampled from the green list is at least:
\[
P[k \in G] \geq \frac{\gamma \alpha}{1+(\alpha-1)\gamma}S(p,\frac{(1-\gamma)(\alpha-1)}{1+(\alpha-1)\gamma}).
\]
\end{lemma}

Therefore, the expected number of green list tokens in a sequence can be expressed as:

\begin{equation}
\mathbb{E}|s|_G \geq \frac{\gamma \alpha T}{1 + (\alpha - 1) \gamma} S^\star,
\label{eq:bound}
\end{equation}

where \( S^\star \) represents the average spike entropy of the sequence, and \( \alpha = \exp(\delta) \).

\paragraph{Theoretical Bound on Perplexity of Watermarked Text.}
\label{ppl}
We introduce an important lemma from \citet{KGW}, which provides the theoretical value of the perplexity of watermarked text:

\begin{lemma}
\label{lemma:low-bound-ppl-kgw}

Consider a sequence \( s(i) \) where \( -N_p < i < T \). Suppose the non-watermarked language model generates a probability vector \( p^{(T)} \) for the token at position \( T \). The watermarked model predicts the token at position \( T \) using a modified probability vector \( \hat{p}^{(T)} \). The expected perplexity of the \( T \)-th token, taking into account the randomness of the red list partition, is given by 

\[
\mathbb{E}_{W,B} \sum_{k} \hat{p}^{(T)}_k \ln(p^{(T)}_k) \leq (1 + (\alpha - 1)\gamma) P^*,
\]

where \( P^* = \sum_{k} p^{(T)}_k \ln(p^{(T)}_k) \) represents the perplexity of the original model. 
\end{lemma}
This tells us that the perplexity of the watermarked text is solely dependent on \( \gamma \) and the additional watermark bias \( \delta \).

\subsection{Example of Inaccurate Non-Watermarked Text Estimations on Detection}
\label{app:example}

\begin{figure*}[t!]
\centering\includegraphics[scale=0.6,trim=10 150 30 70,clip]{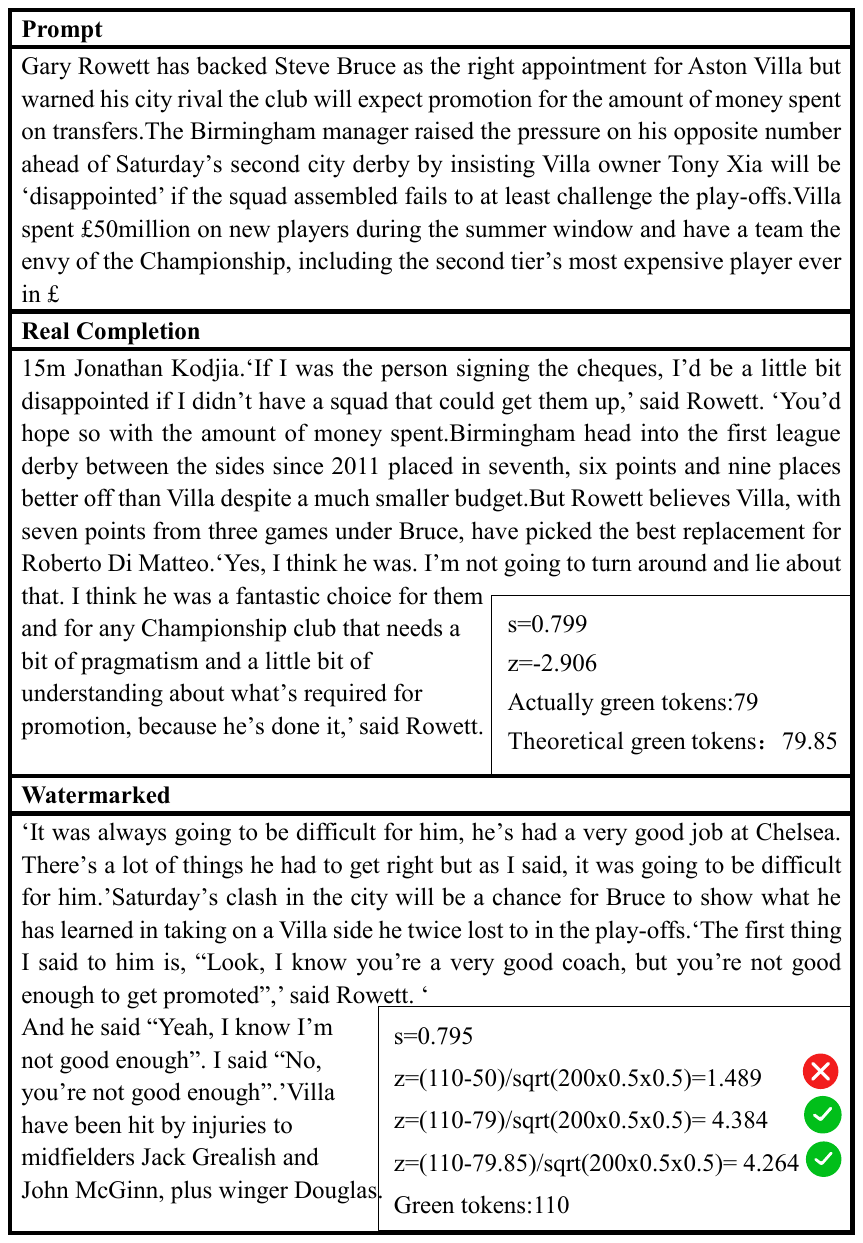}
\caption{Detection Impact from Inaccurate Non-Watermarked Text Estimations with Watermarked Text (where \(\gamma\) is 0.5 and \(\delta\) is 1.0) Generated by OPT-1.3, assuming classification as watermarked text occurs when the \(z\)-value exceeds 4.0.}
\label{fig:example}
\end{figure*}

Figure~\ref{fig:example} illustrates a detection failure caused by inaccurate estimation of green token counts in non-watermarked text, assuming a \(z\)-value threshold of 4.0 for watermark detection. \kgw\ assumes that the green token count in non-watermarked text equals \(\gamma T\), which is not always accurate. In this example, the green token count in human-written text is 79, while the spike entropy values for human-written and watermarked text are 0.799 and 0.795, respectively. According to Equation~\ref{eq:bound}, the theoretical lower bound for green tokens in non-watermarked text is 79.85. Adjusting the expected green token count to 79 or 79.85 enables correct detection of the watermarked text.  

\section{Application of \tool\ in \sweet\ and \ewd}
\label{app:to-sweet-ewd}

\subsection{Applying \tool\ to \sweet}
\sweet\ builds upon \kgw\ and is designed for code generation tasks. Unlike \kgw, \sweet\ applies a bias \( \delta \) to the logits of tokens in \( G \) only if their entropy exceeds a certain threshold $\tau$. In other words, \sweet\ does not enforce the green-red rule on low-entropy tokens.
The formula for calculating the \( z \)-value in \sweet\ is as follows:

\begin{equation}
z = \frac{\left|s\right|^h_G - \gamma T_h}{\sqrt{T_h \gamma (1 - \gamma)}},
\end{equation}

where \(\left|s\right|^h_G\) represents the number of green tokens in \(s\) that have an entropy value \(H_t\) higher than the threshold \(\tau\), and \(T_h\) denotes the total number of tokens that have an entropy value higher than the threshold \(\tau\).

When applying the bipolar watermarking rule, we will only apply a bias \( \delta \) to the logits of tokens in \( G \) for the positive pole or to the logits of tokens in \( R \) for the negative pole, provided that their entropy exceeds the threshold \( \tau \).
The z-statistic for the differential detection method is calculated as follows: 
\begin{equation}
z = \frac{|s|^h_{pG} - |s|^h_{nG}-\gamma T^h_p+(1-\gamma) T^h_n}{\sqrt{T_h\gamma(1-\gamma)}},
\label{eq:z_d_sweet}
\end{equation}

where \(|s|^h_{pG}\) and \(|s|^h_{nG}\) represent the number of positive pole and negative pole green tokens in \(s\) that have an entropy value higher than the threshold \( \tau \). \(T^h_p\) and \(T^h_n\) denote the total number of tokens in \(s\) for the positive and negative poles, respectively, with an entropy value exceeding the threshold \( \tau \).

\subsection{Applying \tool\ to \ewd}


\ewd\ is another watermark detection algorithm based on \kgw. Specifically, it proposes that a token's influence during detection should be proportional to its entropy. To fully reflect this positive relationship between token entropy and detection influence, a monotonically increasing and continuous function is utilized to generate influence weights from token entropy.

The weight \( W(t) \) of a token \( t \) is defined as follows:

\begin{equation}
W(t) = f(SE(t) - C_0),
\label{eq:weight}
\end{equation}

where \( SE(t) \) represents the token's entropy and \( C_0 \) is the minimal value of spike entropy, used to normalize the entropy input before computing the weight.

To calculate the z-score using the provided formula:

\begin{equation}
\label{eq:wed-z}
z = \frac{|s|^W_G - \gamma \sum_{}^{} W_i}{\sqrt{\gamma (1 - \gamma) \sum_{}^{} W_i^2}}.
\end{equation}

EWD modifies only the watermark detection method, 
while the watermark generation algorithm remains the same as \kgw. Therefore, when applying the bipolar watermark, the process as shown in Algorithm~\ref{alg:generator}. 
During detection, we utilize the differential detection algorithm:

\begin{equation}
z = \frac{|s|^W_{pG} -|s|^W_{nG}- \gamma \sum_{i\in p}^{} W_i+(1-\gamma)\sum_{i\in n}^{}W_i}{\sqrt{\gamma (1 - \gamma) \sum_{}^{} W_i^2}},
\end{equation}

where \(|s|^W_{pG}\) and \(|s|^W_{nG}\) represent the weighted sums of green tokens in the positive and negative poles, respectively. Additionally, \(\sum_{i \in p} W_i\) and \(\sum_{i \in n} W_i\) denote the weighted sums of all tokens in the positive and negative poles, respectively.

\section{Proofs of Theorems~\ref{theorem:low-bound-green} and \ref{theorem:no-increase-false-positive}}
\label{app:proof}
\subsection{Proof of Theorem~\ref{theorem:low-bound-green}}
In the \kgw\ framework, we can derive a lower bound for the number of green list tokens in \( s \) by summing the results of lemma~\ref{lemma:low-bound-green-kgw}. 
The lower bound of the expected number of green list tokens is given by:
$\mathbb{E}(|s|_G) \geq \frac{\gamma \alpha T}{1 + (\alpha - 1) \gamma} S^\star $,
where  
\( S^\star \) represents the average spike entropy of the generated sequence.
By applying the definition of the \( z \)-score from Equation~\ref{eq:z_score}, we obtain a lower bound for the \( z \)-score:

\begin{equation}
z_{k} \geq \frac{\frac{\gamma \alpha T}{1 + (\alpha - 1) \gamma}S^\star  - \gamma T}{\sqrt{T \gamma(1 - \gamma)}},
\label{eq:bound_k}
\end{equation}


where \( \alpha = \exp(\delta) \). 

When using bipolar watermarking, the number of green tokens in the positive polarity, according to (\ref{eq:bound}), is at least: 
$\frac{\gamma \alpha T_p}{1 + (\alpha - 1) \gamma} S^\star$,
Similarly, we assume that the entropy of the negative polarity is maximized, meaning that each token follows a uniform probability distribution. Under this condition, the expected number of green tokens in the negative polarity reaches its maximum, given by: 
$\frac{\gamma T_n}{1 + (\alpha - 1) \gamma} $,
the expected difference between the two polarities of green tokens is given by:
\begin{equation}
\label{eq:s_d}
\mathbb{E}|s_d| \geq 
\frac{\alpha \gamma T_p}{1 + (\alpha - 1)\gamma}S_p^\star 
-\frac{\gamma T_n}{1 + (\alpha - 1)\gamma},
\end{equation}

where \( S_p^\star \) is the average spike entropy of the positive polarity sequence.
We substitute this into the calculation formula for the \( z \)-score (\ref{eq:z_d}) to obtain:

\begin{equation}
z_d \geq \frac{\frac{\alpha \gamma T_p}{1 + (\alpha - 1)\gamma}S^\star 
-\frac{\gamma T_n}{1 + (\alpha - 1)\gamma}-\gamma T_p+(1-\gamma)T_n}{\sqrt{T\gamma(1-\gamma)}}.
\label{eq:bound_d}
\end{equation}

We denote the right-hand sides of (\ref{eq:bound_k}) and 
(\ref{eq:bound_d}) as \( \mathcal{B}(|z|_k) \) and \( \mathcal{B}(|z|_d) \), respectively.
The difference \( \mathcal{B}(|z|_d) - \mathcal{B}(|z|_k) \) is given by:

\begin{equation}
\mathcal{B}(|z|_d) - \mathcal{B}(|z|_k) = \frac{\frac{\alpha \gamma T_p}{1 + (\alpha - 1)\gamma}S_p^\star 
-\frac{\gamma T_n}{1 + (\alpha - 1)\gamma}-\gamma T_p+(1-\gamma)T_n}{\sqrt{T\gamma(1-\gamma)}} - \frac{\frac{\gamma \alpha T}{1 + (\alpha - 1) \gamma}S^\star - \gamma T}{\sqrt{T \gamma(1 - \gamma)}}. 
\label{eq:bound_z_d}
\end{equation}

We assume that \( S^\star \) is equal to \( S_p^\star \). By substituting this assumption into Equation (\ref{eq:bound_z_d}), we can simplify it to:

\begin{align}
\mathcal{B}(|z|_d) - \mathcal{B}(|z|_k) &= \frac{-\frac{\gamma \alpha T_n}{1 + (\alpha - 1) \gamma} S^\star - \frac{\gamma T_n}{1 + (\alpha - 1) \gamma}+T_n}{\sqrt{T\gamma(1-\gamma)}} \nonumber \\ 
&\geq \frac{-\frac{\gamma \alpha T_n}{1 + (\alpha - 1) \gamma} - \frac{\gamma T_n}{1 + (\alpha - 1) \gamma}+T_n}{\sqrt{T\gamma(1-\gamma)}}=0,
\end{align}

In the last step, we utilized the fact that the value of cross-entropy cannot exceed 1. Therefore, we have successfully proven Theorem~\ref{theorem:low-bound-green}.

\paragraph{Discussion on the Effectiveness of Applying BiMarker to EWD}

We first introduce the statistical weight and expectation of EWD. Following the original paper~\cite{EWD}, for analysis purposes, we adopt a linear weight function for each token \( k \):  
\[ W(k) = SE(k) - C_0, \]  
where \( C_0 \) is a constant. Based on Lemma~\ref{lemma:low-bound-green-kgw}, the average of the sum of weights for green tokens can be obtained as: 
 $\sum_k P[k \in G] \cdot W(k).$
 By applying the definition of the \( z \)-score from Equation~\ref{eq:wed-z}, we obtain a lower bound for the \( z \)-score:

\begin{equation}
z_{e} \geq \frac{\frac{\gamma \alpha T}{1 + (\alpha - 1) \gamma}SW^\star  - \gamma \sum_{}^{} W_i}{\sqrt{\gamma (1 - \gamma) \sum_{}^{} W_i^2}},
\label{eq:bound_e}
\end{equation}

where \( SW^\star \) represents the mean of the product of entropy and weight.  

Following a similar logic to Equation~(\ref{eq:s_d}) and based on Equation~(\ref{eq:z_d}), the minimum expected value of the detection \( z \)-score, when applying bipolar watermarking and differential detection algorithms, can be derived as:

\begin{equation}
z_d \geq \frac{\frac{\gamma \alpha T_p}{1 + (\alpha - 1) \gamma}SW_p^\star - \frac{\gamma T_p}{1 + (\alpha - 1) \gamma}SW_n^\star - \gamma \sum_{i \in p} W_i + (1 - \gamma) \sum_{i \in n} W_i}{\sqrt{\gamma (1 - \gamma) \sum W_i^2}},
\label{eq:bound_ed}
\end{equation}
where \( SW_p^\star \) and \( SW_n^\star \) denote the mean of the product of entropy and weight for the tokens in the positive and negative poles, respectively. 

We denote the right-hand sides of (\ref{eq:bound_e}) and (\ref{eq:bound_ed}) as \( \mathcal{B}(|z|_e) \) and \( \mathcal{B}(|z|_d) \), respectively. Assuming the mean of the product of entropy and weight for positive and negative samples is identical, the difference \( \mathcal{B}(|z|_d) - \mathcal{B}(|z|_e) \) is expressed as:

\[
\mathcal{B}(|z|_d) - \mathcal{B}(|z|_e) = \frac{-\frac{\gamma \alpha T_n}{1 + (\alpha - 1) \gamma}SW^\star - \frac{\gamma T_n}{1 + (\alpha - 1) \gamma}W_n^\star - \gamma \sum_{i \in p} W_i + (1 - \gamma) \sum_{i \in n} W_i+\gamma \sum W_i}{\sqrt{\gamma (1 - \gamma) \sum W_i^2}}.
\]

Due to the maximum value of spiked entropy being less than 1, we simplify the numerator and obtain:
\[
\mathcal{B}(|z|_d) - \mathcal{B}(|z|_e) \geq \frac{T_n W^\star -\sum_{i \in n} W_i}{\sqrt{\gamma (1 - \gamma) \sum W_i^2}} = 0.
\]

\subsection{Proof of Theorem~\ref{theorem:no-increase-false-positive}}

We assume that when the \( z \) value exceeds the threshold \( z_{\text{threshold}} \), the text being tested is classified as watermarked. We define \( t = z_{\text{threshold}} \sqrt{T \gamma (1 - \gamma)} \). Thus, the false positive rate when using the \kgw\ detection method for non-watermarked text can be expressed as:

\begin{equation}
F_{\text{KGW}} = P(x > \mu_T+t | \text{non-watermarked})=\int_{\mu_T+t}^{+\infty} \frac{1}{\sqrt{2\pi}\sigma_T} e^{-\frac{(x-\mu_T)^2}{2\sigma_T^2}} \, dx=1 - \Phi(\frac{t}{\sigma_T}).
\end{equation}

where \( x \) represents the number of green tokens in the non-watermarked text,
\(\Phi\) represents the cumulative distribution function of the standard normal distribution.

Under the differential method, if \( y \) and \( x \) are the numbers of green tokens in the positive and negative polarities, respectively, and \( y - x > t \), the text is classified as watermarked. The false positive rate can then be expressed as:  

\begin{equation}
F_{\text{Diff}} = P(y - x > t \mid \text{non-watermarked}) = \int_{-\infty}^{+\infty} \int_{x + t}^{+\infty} f(x) f(y) \, dy \, dx=\int_{-\infty}^{+\infty} 1 - \Phi(\frac{x+t-\mu_{T_n}}{\sigma_{T_p}})\, dx
\label{eq:fp2}
\end{equation}

where \( f(x) \) and \( f(y) \) are the Gaussian probability density functions for the negative and positive polarities, respectively:  

\[
f(x) = \frac{1}{\sqrt{2\pi} \sigma_{T_n}} e^{-\frac{(x - \mu_{T_n})^2}{2 \sigma_{T_n}^2}}, \quad 
f(y) = \frac{1}{\sqrt{2\pi} \sigma_{T_p}} e^{-\frac{(y - \mu_{T_p})^2}{2 \sigma_{T_p}^2}}.
\]

Continuing to simplify (\ref{eq:fp2}), 
since the \(\Phi\) function is monotonically 
increasing and \( x - \gamma T_n > 0 \), along with \(\sigma_{T_p} \leq \sigma_T\), we can derive:  

\begin{equation}
F_{\text{Diff}} \leq \int_{-\infty}^{+\infty} 1 - \Phi(\frac{t}{\sigma_T}) \, dx= 1 - \Phi(\frac{t}{\sigma_T})=F_{KGW}.
\end{equation}

Thus, Theorem~\ref{theorem:no-increase-false-positive} is proven.



\paragraph{Discussion on the Effectiveness of Applying BiMarker to EWD.}

Our proof process is highly similar to that of \kgw. The key difference lies in \( t = z_{\text{threshold}} \sqrt{\gamma (1 - \gamma) \sum W_i^2} \), and the Gaussian distribution represents the distribution of green token weights rather than the distribution of the number of green tokens.

\section{Experimental Details}

\subsection{Hyper-parameters}
\label{app:experiment-setting}

\paragraph{Hyper-parameters.}
For high-entropy tasks, 
multinomial sampling employs a fixed sampling temperature of 0.7.
In contrast, beam-search sampling utilizes 8 beams, with a $no\_repeated\_ngrams$ 
constraint set to 16 to mitigate excessive text repetition. For low-entropy tasks, specifically code generation, we adopt top-p sampling~\cite{top-p} with \( p = 0.95 \) and a lower temperature of 0.2.
When using \sweet\ for watermark embedding and detection, we set the entropy threshold to 0.695, consistent with~\cite{sweet,EWD}, and exclude all tokens below this threshold during both processes.

\subsection{Evaluating the Impact of \tool\ on Text Quality}
\label{app:curve}
\begin{figure*}[ht]
\vskip 0.2in
\begin{center}
\centerline{\includegraphics[width=1\columnwidth]{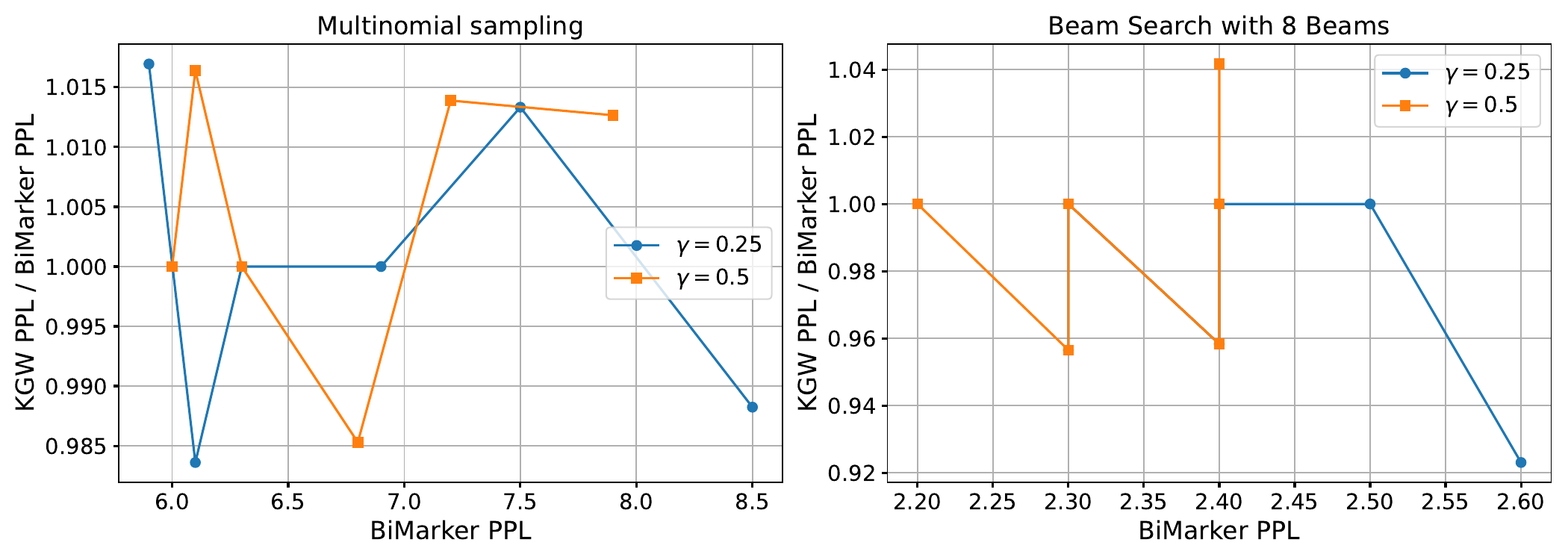}}
\caption{PPL comparison under different watermark strengths ($\delta \in [0.5, 0.75, 1.0, 1.5, 2, 2.5]$), where the x-axis denotes \tool’s PPL and the y-axis shows the ratio of \kgw’s PPL to \tool’s.
}
\label{fig:ppl}
\end{center}
\vskip -0.2in
\end{figure*}

Figure~\ref{fig:ppl} presents a PPL comparison under 
different watermark strengths,
where the x-axis denotes \tool’s PPL and the y-axis shows the ratio of \kgw’s PPL to \tool’s. We observe that, under the same watermark strength, \tool\ and \kgw\ exhibit similar PPL values, indicating that \tool\ does not significantly alter the PPL compared to \tool, which aligns with our theoretical prediction.

Table~\ref{tab:pass1} presents the impact of applying \tool\ to various watermark embedding algorithms on the quality of generated code. 
Note that EWD does not include a generation algorithm. 
The results clearly indicate that \tool\ leaves code generation quality essentially unchanged.

\begin{table}[t]
\caption{The accuracy (pass@1) of the generated code using different methods($\gamma=0.25, \delta=2$).}
\label{tab:pass1}
\vskip 0.15in
\begin{center}
\begin{small}
\begin{sc}
\begin{tabular}{lcc}
\toprule
 methods  & humaneval & mbpp \\
\midrule
  \kgw\    & 0.274  &0.326\\
\kgw +\tool\     & 0.287  &0.336\\ \cline{1-3}
  \sweet\    & 0.311  &0.334      \\
  \sweet+\tool\ & 0.323  &0.31 \\ 
\bottomrule
\end{tabular}
\end{sc}
\end{small}
\end{center}
\vskip -0.1in
\end{table}

\end{document}